\documentclass{article}
\usepackage[final]{nips_2017}
\usepackage{graphicx,subfigure} 
\usepackage{amsmath,amssymb}
\usepackage{algorithm}
\usepackage{algorithmic}
\input{def.set}

\begin{document}
\title{Bayesian Semi-nonnegative Tri-matrix Factorization to Identify Pathways Associated with Cancer Types
} 

\author{
Sunho Park\\
Department of Quantitative Health Sciences\\
Cleveland Clinic, Cleveland, OH, 44195\\
\texttt{ parks@ccf.org}
\And
Tae Hyun Hwang\\
Department of Quantitative Health Sciences\\
Cleveland Clinic, Cleveland, OH, 44195\\
\texttt{ hwangt@ccf.org}
}

%
\newcommand{\fix}{\marginpar{FIX}}
\newcommand{\new}{\marginpar{NEW}}
\maketitle

\begin{abstract}
Identifying altered pathways that are associated with specific cancer types can potentially bring a significant impact on cancer patient treatment. Accurate identification of such key altered pathways information can be used to develop novel therapeutic agents as well as to understand the molecular mechanisms of various types of cancers better. Tri-matrix factorization is an efficient tool to learn associations between two different entities (e.g., cancer types and pathways in our case) from data. To successfully apply tri-matrix factorization methods to biomedical problems, biological prior knowledge such as pathway databases or protein-protein interaction (PPI) networks, should be taken into account in the factorization model. However, it is not straightforward in the Bayesian setting even though Bayesian methods are more appealing than point estimate methods, such as a maximum likelihood or a maximum posterior method, in the sense that they calculate distributions over variables and are robust against overfitting. We propose a Bayesian (semi-)nonnegative matrix factorization model for human cancer genomic data, where the biological prior knowledge represented by a pathway database and a PPI network is taken into account in the factorization model through a finite dependent Beta-Bernoulli prior. We tested our method on The Cancer Genome Atlas (TCGA) dataset and found that the pathways identified by our method can be used as a prognostic biomarkers for patient subgroup identification.
\end{abstract}

\vspace{-0.7cm}
\section{Introduction}
\label{sec:intro}
\vspace{-0.5cm}

Matrix factorization and its various models (including bi- and tri-matrix factorizations) that utilize genomic data have been wildly used to identify key altered pathways\footnote{Pathway is defined as "a series of actions among molecules in a cell that leads to a certain product or a change in the cell." (quoted from https://www.genome.gov/27530687/biological-pathways-fact-sheet/). In our problem setting, a pathway simply means a set of genes that lead a certain biological function.} associated with cancer development, progress, and patient subgroup identification in cancer genomics.
One of key challenges is how to elaborately include biological prior knowledge, e.g., a pathway database or a protein-protein interaction (PPI) network, into the factorization models. However, most approaches are based on regularization approaches, where additional terms that encourage a solution to be similar to what we expect from the prior knowledge are introduced into an objective function. For example, consider the case where prior knowledge is given in the form of a PPI network. We expect that parameters associated with two genes that are connected in the PPI network should have similar values to each other. We can take this smoothness constraint into account by adding a regularizer term involving a Laplace matrix constructed from the PPI network to an objective function (e.g., [7]). However, this approach yields just a point estimate: the uncertainty of the solution is neglected and the methods often suffer from overfitting. Bayesian approaches are more appealing because they calculate a full distribution over parameters, instead of a single point estimate. However, it is not straightforward to take the prior knowledge into account in a factorization model in the Bayesian framework.

We here propose a Bayesian (semi-)nonnegative matrix factorization for human cancer genomic data,
where the biological prior knowledge represented by a pathway database and a PPI network is incorporated into
the factorization through a finite dependent Beta-Bernoulli prior model. We tested our method on The Cancer Genome Atlas (TCGA) dataset and found that the pathways identified by our method can be used as a prognostic biomarkers for patient subgroup identification.

\vspace{-0.2cm}
\textbf{Notations}: For a matrix $\bA$, $\ba_i$ represents its $i$th row vector, i.e., $\ba_i \triangleq (\bA_{i,:})^{\top}$. Similarly, $\vec{\ba}_j \triangleq \bA_{:,j}$ refers to its $j$th column vector. The $(i,j)$th element of the matrix $\bA$ is expressed by $A_{ij}$.

\vspace{-0.3cm}
\section{Related work}
\vspace{-0.4cm}
We assume that observations are given in a form of matrix, $\bX \in \Real^{N \times D}$, where N and D are the number of samples and input features respectively. The goal of tri-matrix factorization is to approximate the matrix $\bX$ by a multiplication of three sub latent matrices, such that
\vspace{-0.1cm}
\be
    \label{eq:factorization_simple}
    \bX \approx \bU \bS \bV^\top,
\ee
where $\bU \in \Real^{N \times K}$, $\bS \in \Real^{K \times R}$ and $\bV \in \Real^{R \times D}$. Here,
$K$ and $R$ are set to smaller numbers compared to $N$ and $D$.
One can add further constraints based on the properties of the observations and available prior knowledge. In fact, most of previous work assumes that the observation matrix is non-negative
and finds all non-negative sub latent matrices [4, 10].
However, in this paper, we relax this restriction.
Motivated by [3], we assume that $\bX$ is real-valued and that the sub matrices, $\bU$ and $\bS$, are still non-negative but $\bV$ is real-valued, which will be explained in detail in the next section.

Nonnegative matrix factorization methods, including the bi-matrix and the tri-matrix factorization models, have been applied to many different biological problems (please see references in [2]). In particular, in [7] the authors use the nonnegative tri-matrix factorization model to identify pathways that are relevant to human cancer, which is the same goal as ours.
The method also involves a pathway database and a PPI network to learn the latent matrices
in the regularization approach. However, this approach gives us just a single point estimate.
Furthermore, the regularized coefficients should be pre-specified (tuned) by users.
For large-scale data problems, it requires heavy-computational burden when we find optimal values for regularized coefficients via cross-validation.

A Bayesian nonnegative tri-matrix factorization approach has been proposed [1],
where exponential prior distributions are placed on nonnegative latent matrices and their posterior distributions are calculated by Gibbs sampling or the variational inference.
However, the method does not consider how to use available biological prior knowledge to learn the sub-matrices in the factorization model.

\vspace{-0.3cm}
\section{Bayesian (semi-)nonnegative matrix tri-factorization}
\vspace{-0.4cm}
We propose a Bayesian (semi-)nonnegative matrix factorization for human cancer genomic data,
where the prior knowledge represented by a pathway database and a PPI network
is taken into account in the factorization through a finite dependent Beta-Bernoulli prior model.
One can consider our method as an extension of the work in [7] in the Bayesian framework. Thus, our method shares several nice properties inherited from Bayesian approaches.
Furthermore, unlike the work in [7],
we use gene expression data, instead of mutation data (since it usually yields a highly sparse observation matrix). For gene expression data, the observation matrix $\bX$ includes negative values (it is same for other types of genomic data, such as copy number, miRNA expression etc).
As mentioned earlier, we extend the concept of the semi-negative factorization in [3] to the tri-matrix factorization case, where a certain latent matrix is allowed to have negative values. However, our method can be applied to nonnegative cases as well with minor changes (placing nonnegativity priors on all latent matrices).

Now, we assume the matrix $\bX$ is gene expression data measured from patients: the $(i,j)$th element in $\bX$, $X_{ij}$, represents an expression value at the $j$th gene from the $i$th patient sample. Furthermore, we construct $\bU$ from patient-cancer-type information:
$K$ is the number of known cancer types and $U_{ij}=1$ indicates the $i$th patient has $j$th cancer type.
We assume that each patient has a single cancer type, i.e., $\sum_{k=1}^{K}U_{ik} = 1$ and $U_{ik} \in \{ 0,1\}$. In addition, let denote $\bZ^0\in\Real^{D \times R}$ by a given pathway database where $R$ is the number of pathways in the database and $Z^{0}_{jr}=1$ if the $j$th gene is annotated in $r$th pathway as a member. With the binary matrix $\bZ^{0}$, our factorization model is:
\vspace{-0.1cm}
\be
    \label{eq:factorization}
    \bX \approx \bU \bS \parenbig{\bZ^0 \circ\bV}^\top
\ee
where $\circ$ is Harmardard operator, i.e., element-wise multiplication operator.
In our model (\ref{eq:factorization}), the matrix $\bV$ is a set of basis (row) vectors and
only few elements (corresponding to member genes of each pathway) in the matrix can contribute to the factorization due to the element-wise multiplication with the binary matrix $\bZ^0$.
Then, the matrix $\bS$ represents associations between cancer types and pathways, where each element of $S_{ij}$ represents associations between $i$th cancer type with $j$th pathway. Higher values of elements indicate stronger associations between cancer types and pathways. Thus, our goal can be satisfied by finding an accurate association matrix $\bS$ from the data.

We next need to specify the likelihood model. We assume it to follow a Gaussian distribution:
\be
    X_{ij} \sim \calN(X_{ij} | \bu_i^\top \bS \parenbig{\bz_j \circ \bv_j}, \gamma),   \qquad \gamma \sim \mbox{Gam}(\gamma|\alpha_{a}^0,\alpha_{b}^0)
\ee
where the precision $\gamma$ (the inverse of the variance) is sampled from a Gamma distribution.

Additional assumptions on the latent matrices are made as follows.
With the fixed $\bU$, the reconstructed matrix, i.e., $\hat{\bX}=\bU \bS \parenbig{\bZ^0 \circ\bV}^\top$ will have the same valued rows for all the samples that belong to the same cancer types.
To remedy this, we also learn the cluster matrix $\bU$ from the data:
\be
    \bu_i = \zeta\bu^0_i + (1-\zeta) \widetilde{\bu}_i,
\ee
where $\bu^0_i$ is directly from the patient-cancer-type data, $\zeta>0$ and $\widetilde{\bu}_i \in \Real^{K}$ is also a probability vector to be learned, i.e., $\sum_{k=1}^{K}\widetilde{U}_{ik} = 1$ and $U_{ik} \ge 0$. The closer $\zeta$ to 1, the more we rely on the given information.
Next, an element in $\bS$ is sampled from an Exponential distribution (nonnegativity):
\be
    S_{kr} \sim \mbox{Expon}(S_{kr}| \lambda^{S0}_{kr}).
\ee
Furthermore, as mentioned earlier, since the observation matrix is real valued, we allow $\bV$ to have non-negative values. We assume each element in $\bV$ to be sampled from a Gaussian distribution.
\be
    V_{jr} \sim \calN(V_{jr}|\mu_{jr}^{V0},\sigma_{jr}^{V0}).
\ee

Since a pathway database could be incomplete, we also learn a binary matrix $\bZ\in \Real^{D\times R}$ based on $\bZ^0$ with a PPI network (in a form of a graph, $\calG$). To do this, we place a finite dependent Beta-Bernoulli prior distribution on $\bZ$ [9]. For better understating, we first consider a finite Beta-Bernoulli prior distribution without considering the dependency, which is given in a hierarchical way:
\be
    Z_{jr} \sim \mbox{Bernoulli\,n}(\pi_{r}), \quad \pi_{r} \sim \mbox{Beta}(\beta_a/R,1),
\ee
The expected number of non-zero entries in $\bZ$ is $D\beta_a/(1 + \beta_a/R)$ and this parametric (finite) model becomes Indian Buffet process when R goes to infinity [5].
We can naturally think that two genes may work together if they are connected in $\calG$. In other words, two connected genes in the graph would be active together in a pathway. First, let us define $\bG \in \Real^{D\times R}$ as a matrix containing $R$ sets of function values and assume that each column, $\vec{\bg}_{r}$, follows a Gaussian process given by
\be
    \vec{\bg}_{r}|\calG \sim \calN(\vec{\bg}_{r}| 0, \bL)
\ee
where $\bL = \bI - \bD^{-1/2}\bA\bD^{-1/2}$ is a normalized Laplacian matrix. we now couple a Bernoulli variable $Z_{jr}$ and a function value $G_{jr}$ using the same method in [8, 9]:
\be
    Z_{jr} =  \mathbb I\Big[G_{jr}<\Phi^{-1}(\pi_r)\Big],\quad
    G_{jr} \sim \calN\big(0,1\big), \quad \pi_r \sim \mbox{Beta}\big({\beta_a}/{R}, 1\big),
\ee
where $\Phi$ is a cumulative Gaussian distribution.
For example, $Z_{jr}$ and $Z_{j'r}$ are more likely to be both 1 or 0 if $G_{jr}$ and $G_{j'r}$ are close to each other. For simplicity, we define $\bar{\pi}_r = \Phi^{-1}(\pi_r)$:
\be
    p(\bar{\pi}_r) = \mbox{Beta}\big(\Phi(\bar{\pi}_r)|\beta_a/R,1 \big)\calN(\bar{\pi}_r|0,1).
\ee
Lastly, we define $\calM = \{ (j,r)| Z^0_{jr} =1 \}$ as nonzero entries in $\bZ^0$ and try to make sure that all the corresponding entries in $\{Z_{jr}| (j,r)\in\calM \}$ are also close to 1 during training.

\vspace{-0.3cm}
\section{Variational inference}
\vspace{-0.4cm}
We approximately compute posterior distributions of all the variables in the variational inference framework.
First, the variational distributions are assumed to be fully factorized:
\be
q(\gamma,\bS,\bZ,\bV,\bg,\bpi) =
        q(\gamma)\Big(\prod_{k=1}^K\prod_{r=1}^R q(S_{kr})\Big)\Big(\prod_{j=1}^D\prod_{r=1}^R q(V_{jr})q(Z_{jr}) q(G_{jr}) \Big)
        \Big( \prod_{r=1}^R q(\bar{\pi}_r) \Big).
\ee
The form of each variational distribution is as follows
\be
    & & q(\gamma) = \mbox{Gam}(\gamma|\alpha_{a},\alpha_{b}), \quad q(S_{kr}) = \calT\calN (S_{kr}|\mu^{S}_{kr},\sigma^{S}_{kr}) , \quad q(V_{jr}) = \calN (V_{jr}|\mu^{V}_{jr},\sigma^{V}_{jr}), \nonumber \\
    & & q(Z_{jr}) = p(Z_{jr}|\pi_{r},G_{jr}), \quad q(G_{jr}) = \calN (G_{jr}|\mu^{g}_{jr},\sigma^{g}_{jr}) , \quad q(\bar{\pi}_{r}) = \calN (\bar{\pi}_{r}|\mu^{\bar{\pi}}_{r},\sigma^{\bar{\pi}}_{r}).
\ee
Note that $q(Z_{jr}$) is fully determined by $\bar{\pi}$ and $G_{jr}$ (also refer (\ref{eq:marginal_q_Z})). Denote a set of all the variables and the parameters by $\Theta = \{\gamma,\widetilde{\bU},\bS,\bV,\bG,\bar{\bpi}\}$.
Then the variational distributions can be computed by maximizing the lower bound on the likelihood
\be
    & & \max_{q(\Theta)} \calL(q) = \int q(\Theta) \log \frac{p(\bX, \Theta)}{q(\Theta)} d\Theta \\
    & & \mbox{such that} \quad q(Z_{jr}=1)=1, \,\, \forall \,\, (j,r)\in \calM,
\ee
where a marginal distribution of an Bernoulli variable $Z_{jr}$, $q(Z_{jr}=1)$, can be calculated as follows
\be
    \label{eq:marginal_q_Z}
    q(Z_{jr}=1) =  \ave_{p(Z_{jr}|g_{jr}\pi_r)q(G_{jr})q(\bar{\pi}_r)}{\Big[Z_{jr} \Big]}
                =  \Phi\Big( {\big(\mu_r^{\bar{\pi}}-\mu^{g}_{jr}\big)}/{\sqrt{\sigma^{\bar{\pi}}_r + \sigma_{jr}^g}} \Big)
\ee
The variational distributions, $q(\gamma)$, $\{q(S_{kr})\}$ and $\{q(V_{jr})\}$, can be updated in closed form (by the similar manner as in [1]). With defining new variables, $\widetilde{U}_{ik}={\exp(\theta_{ik}^U)}/{\sum_{k'=1}^K \exp(\theta_{ik'}^U)}$, the matrix $\widetilde{\bU}$ (obtained by optimizing $\{\theta_{ik}^U\}$) can be found by any unconstrained gradient methods. For $q(\bar{\bpi})$ and $q(\bG)$, we solve the following regularization problem,
\be
    & & \max_{q(\bar{\bpi}),q(\bG)} \calL(q) + \xi\sum_{(j,r) \in \calM}\log q(Z_{jr}=1)
\ee
where $\xi>0$ is a regularization coefficient. Here, the regularization function (a sum of the negative cross entropy) is defined on the posterior distributions over $\{Z_{jr}|(j,r) \in \calM\}$ [11].

\vspace{-0.3cm}
\section{Experimental results}
\vspace{-0.4cm}
We tested our method on the gene expression (mRNA) data from TCGA\footnote{The data was downloaded from CBioportal (http://www.cbioportal.org/). The downloading option was 'TCGA\_cancer\_type\_rna\_seq\_v2\_mrna'.}. We further removed several cancer types that have small number patient samples (less than 200). The number of cancer types and patient samples were $K=20$ and $N=8,444$, respectively. We used 'biocarta' pathway database\footnote{ the data is available from http://software.broadinstitute.org/gsea/msigdb/collections.jsp} where the number of pathways is $R=217$. We also downloaded a PPI network from BioGRID\footnote{https://thebiogrid.org/.
The version is {BIOGRID-ORGANISM-Homo\_sapiens-3.4.153}}. After intersecting all the datasets,
the number of the common genes was $D=1,152$.

\begin{figure}[ht]
\begin{center}
\centerline{
\includegraphics[width=2in]{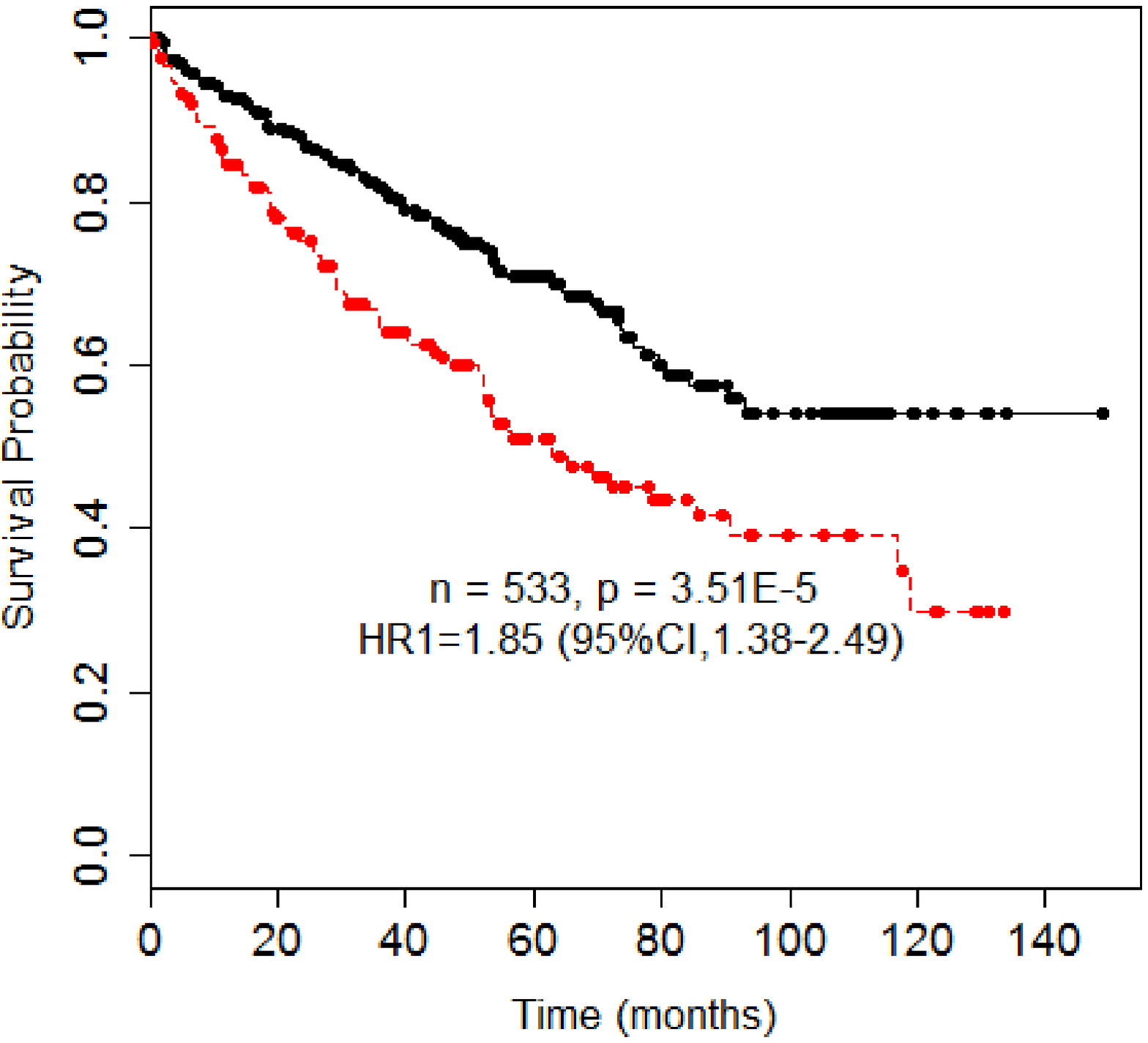}
\hspace{1cm}
\includegraphics[width=2in]{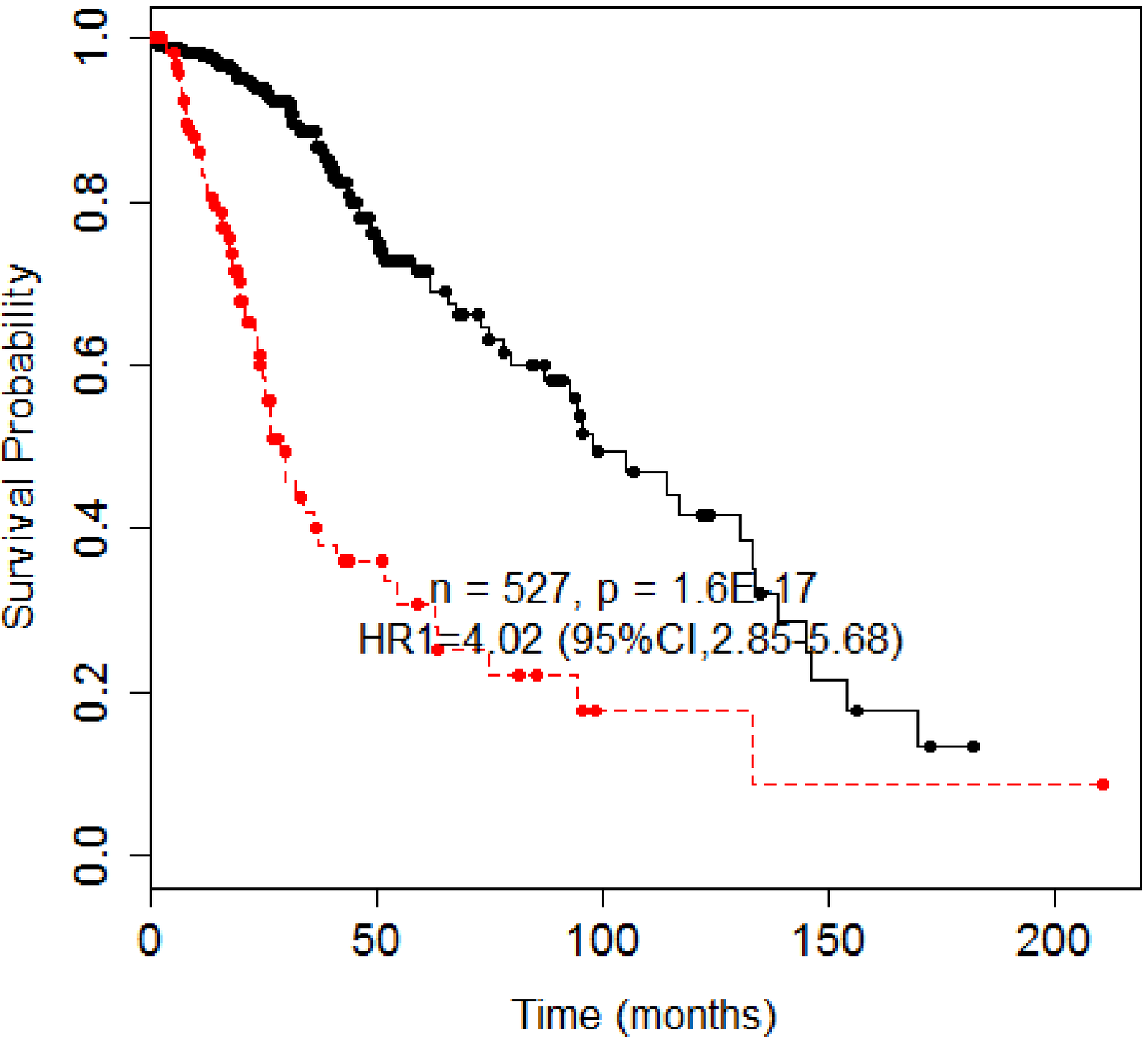}}
\centerline{(a) \hspace{5cm} (b)}
\caption{Kaplan-Meier plot for (a) KIRC (log-rank p-val = $3.51e^{-5}$) and LGG (p-val = $1.60e^{-17}$).}
\label{fig:kmplot}
\end{center}
\end{figure}

After all the parameters (including the latent matrices) were learned, we selected the top-5 ranked pathways for the $k$th cancer type based on the mean values of the elements in the $k$th row vector of $\bS$. All the member genes in these pathways were used for gene signature (biomakers) for the patient stratification of that cancer type.
For each cancer type, we performed consensus clustering on the TCGA gene expression again (500 K-means repetition with bootstrapping [6]) and finally reported two representative results. As shown in Figure \ref{fig:kmplot}, there is a clear separation between the two groups, which suggests that the pathways identified by our method can be used as prognostic biomarkers.

\vspace{-0.5cm}

\section*{References}
[1] Thomas Brouwer and Pietro Lio’. Fast bayesian non-negative matrix factorisation and trifactorisation.
In NIPS 2016 Workshop: Advances in Approximate Bayesian Inference, 2016.

\vspace{0.1cm}
[2] Karthik Devarajan. Nonnegative matrix factorization: An analytical and interpretive tool in
computational biology. PLoS Computational Biology, 4, 2008.

\vspace{0.1cm}
[3] C. Ding, T. Li, and M. I. Jordan. Convex and semi-nonnegative matrix factorizations. Technical
Report 60428, Lawrence Berkeley National Lab, 2006.

\vspace{0.1cm}
[4] C. Ding, T. Li, W. Peng, and H. Park. Orthogonal nonnegative matrix tri-factorizations for
clustering. In Proceedings of the ACM SIGKDD Conference on Knowledge Discovery and
Data Mining (KDD), Philadelphia, PA, 2006.

\vspace{0.1cm}
[5] T. L. Griffiths and Z. Ghahramani. The Indian buffet process: An introduction and review.
Journal of Machine Learning Research, 12:1185–1224, 2011.

\vspace{0.1cm}
[6] Stefano Monti, Pablo Tamayo, Jill Mesirov, and Todd Golub. Consensus clustering: A
resampling-based method for class discovery and visualization of gene expression microarray
data. Machine Learning, 52(1):91–118, 2003.

\vspace{0.1cm}
[7] Sunho Park, Seung-Jun Kim, Donghyeon Yu, Samuel Pea-Llopis, Jianjiong Gao, Jin Suk Park,
Beibei Chen, Jessie Norris, Xinlei Wang, Min Chen, Minsoo Kim, Jeongsik Yong, Zabi Wardak,
Kevin Choe, Michael Story, Timothy Starr, Jae-Ho Cheong, and Tae Hyun Hwang. An integrative
somatic mutation analysis to identify pathways linked with survival outcomes across
19 cancer types. Bioinformatics, 32(11):1643–1651, 2016.

\vspace{0.1cm}
[8] Erik B. Sudderth and Michael I. Jordan. Shared segmentation of natural scenes using dependent
pitman-yor processes. In Advances in Neural Information Processing Systems (NIPS), 2009.

\vspace{0.1cm}
[9] Sinead Williamson, Peter Orbanz, and Zoubin Ghahramani. Dependent indian buffet processes.
In Proceedings of the International Conference on Artificial Intelligence and Statistics
(AISTATS), 2010,.

\vspace{0.1cm}
[10] J. Yoo and S. Choi. Probabilistic matrix tri-factorization. In Proceedings of the IEEE International
Conference on Acoustics, Speech, and Signal Processing (ICASSP), Taipei, Taiwan,
2009.

\vspace{0.1cm}
[11] Jun Zhu, Ning Chen, and Eric P. Xing. Bayesian inference with posterior regularization and
applications to infinite latent svms. Journal of Machine Learning Research, 15(1), 2014.

\end{document}